\documentclass[sigconf]{acmart}

\usepackage{booktabs} % For formal tables
\usepackage{todonotes}
\usepackage{subcaption}
\usepackage{placeins}
\usepackage{afterpage}
\usepackage[utf8]{inputenc}

\setcopyright{none}
\acmDOI{xx.xxx/xxx_x}

\acmConference[Preprint]{Preprint}{March 2019}
\begin{document}
\title{Fashion Outfit Generation for E-commerce}
%\titlenote{Produces the permission block, and
%  copyright information}

\author{Elaine M. Bettaney}
\orcid{1234-5678-9012}
\affiliation{%
  \institution{ASOS.com}
  \city{London}
  \country{UK}
}
\email{elaine.bettaney@asos.com}

\author{Stephen R. Hardwick}
\affiliation{%
  \institution{ASOS.com}
  \city{London}
  \country{UK}
}
\email{stephen.hardwick@asos.com}

\author{Odysseas Zisimopoulos}
\affiliation{%
 \institution{ASOS.com}
 \city{London}
 \country{UK}
 }
 \email{odysseas.zisimopoulos@asos.com}
 
\author{Benjamin Paul Chamberlain}
\affiliation{%
  \institution{ASOS.com}
  \city{London}
  \country{UK}
}
 \email{ben.chamberlain@asos.com}

\renewcommand{\shortauthors}{E. M. Bettaney et al.}
\renewcommand{\vec}[1]{\mathbf{#1}}%

\settopmatter{printacmref=false}

\begin{abstract}
Combining items of clothing into an outfit is a major task in fashion retail. Recommending sets of items that are compatible with a particular seed item is useful for providing users with guidance and inspiration, but is currently a manual process that requires expert stylists and is therefore not scalable or easy to personalise. We use a multilayer neural network fed by visual and textual features to learn embeddings of items in a latent style space such that compatible items of different types are embedded close to one another. We train our model using the ASOS outfits dataset, which consists of a large number of outfits created by professional stylists and which we release to the research community. Our model shows strong performance in an offline outfit compatibility prediction task. We use our model to generate outfits and for the first time in this field perform an AB test, comparing our generated outfits to those produced by a baseline model which matches appropriate product types but uses no information on style. Users approved of outfits generated by our model 21\% and 34\% more frequently than those generated by the baseline model for womenswear and menswear respectively.
\end{abstract}


\maketitle

% Standardise the references: capitalisation, author initialse, pages, conference / journal format

\section{Introduction}
\label{sec:intro}
% Generally should cover 
% - background about ASOS 
% - challenges of outfit generation (different to cross-sell or up-sell)
% - why does the fashion industry need outfit generation
% - other companies doing outfit generation
% - our contribution

% Outfit composition
% Outfit composition is a common problem in fashion. 

User needs based around outfits include answering questions such as "What trousers will go with this shirt?",  "What can I wear to a party?" or "Which items should I add to my wardrobe for summer?".  The key to answering these questions requires an understanding of \emph{style}. Style encompasses a broad range of properties including but not limited to, colour, shape, pattern and fabric.  It may also incorporate current fashion trends, user's style preferences and an awareness of the context in which the outfits will be worn.  In the growing world of fashion e-commerce it is becoming increasingly important to be able to fulfill these needs in a way that is scalable, automated and ultimately personalised.

%\todo[inline]{Consider removing the next para. It's overly commercial and the obvious question is why do asos not already have one? Are we behind the research curve?}
% Other companies doing outfits
%Many online fashion companies are already focusing on this area.  Amazon have launched \emph{Outfit Compare}, a service where the user uploads two photos of themselves wearing different outfits and will be told %which outfit looks the best. Zalando offers a \emph{Get the Look} section which showcases outfits worn by fashion influencers.  Startups, Intelistyle and Chicisimo allow users to upload photos from their own %wardrobe and suggest products to go with them.  Stitch Fix and Thread instead aim to learn a user's style and give personalised recommendations both for products and outfits.

\begin{figure}
    \centering
    \includegraphics[width=0.45\textwidth]{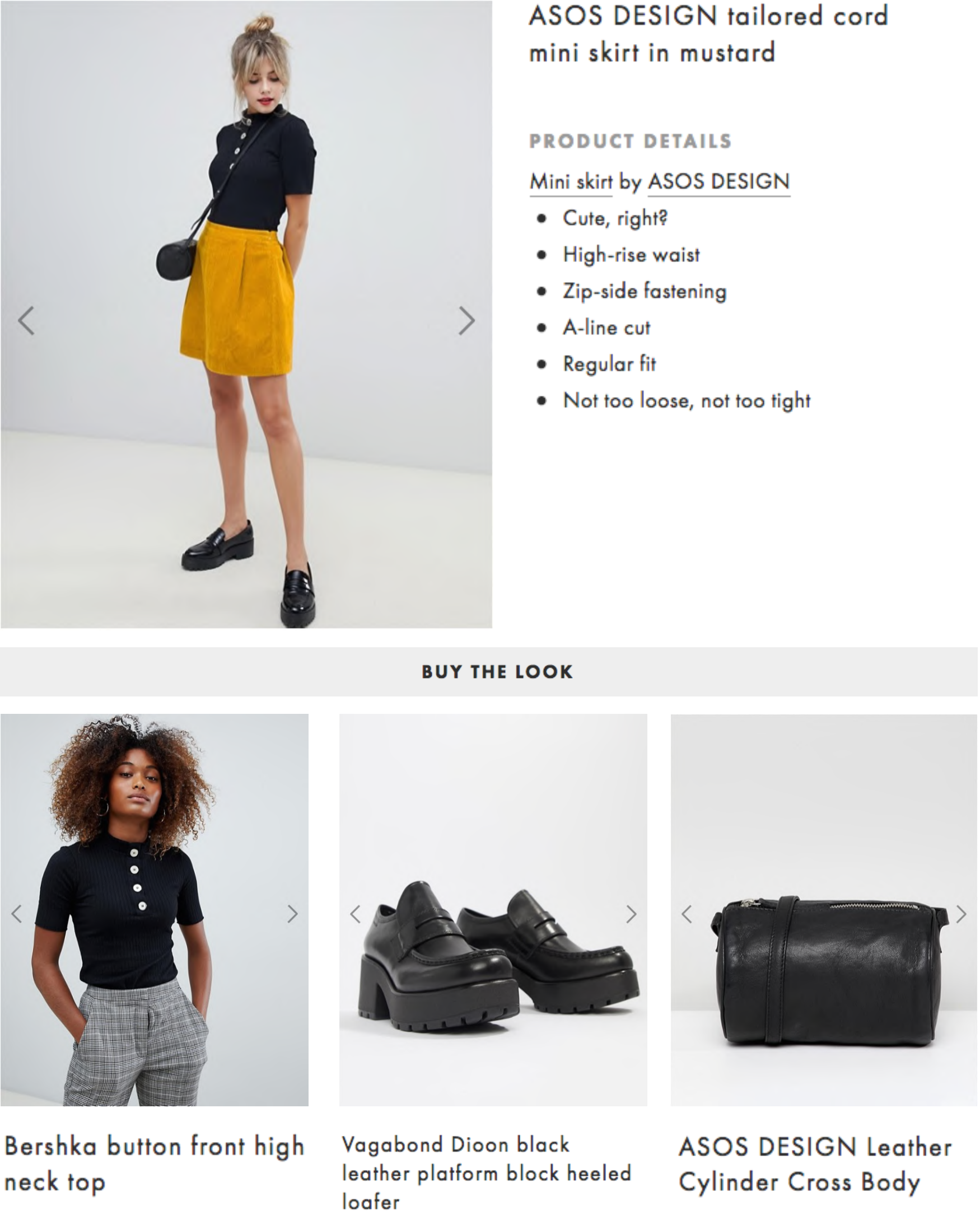}
    \caption{An ASOS fashion product together with associated product data and styling products in a \emph{Buy the Look} (BTL) carousel as shown on a Product Description Page (PDP).}
    \label{fig:buythelook}
    \vspace{-2mm}
\end{figure}

% ASOS
 This paper describes a system for Generating Outfit Recommendations from Deep Networks (GORDN) under development at ASOS.com.  ASOS is a global e-commerce company focusing on fashion and beauty.  With approximately 87,000 products on site at any one time, it is difficult for customers to perform an exhaustive search to find products that can be worn together. Each fashion product added to our catalogue is photographed on a model as part of an individually curated outfit of compatible products chosen by our stylists to create images for its Product Description Page (PDP). The products comprising the outfit are then displayed to the customer in a \emph{Buy the Look} (BTL) carousel (Figure \ref{fig:buythelook}).  This offering however is not scalable as it requires manual input for every outfit.  We aim to learn from the information encoded in these outfits to automatically generate an unlimited number of outfits.

% Usecase and dataset
%\todo[inline]{Ben: I think this is more general. I believe most people get dressed this way too} 
A common way for people to compose outfits is to first pick a seed item, such as a patterned shirt, and then find other compatible items. We focus on this task: completing an outfit based on a seed item.  This is useful in an e-commerce setting as outfit suggestions can be seeded with a particular product page or a user's past purchases. Our ASOS outfits dataset comprises a set of outfits originating from BTL carousels on PDPs. These contain a seed, or `hero product', which can be bought from the PDP. All other items in the outfit we refer to as `styling products'.

% Asymmetry between hero and styling products
There is an asymmetry between hero and styling products.  Whilst all items are used as hero products (in an e-commerce setting), styling products are selected as the best matches for the hero product and this matching is directional. For example when the hero product is a pair of Wellington boots it may create an engaging outfit to style them with a dress. However if the hero product is a dress then it is unlikely a pair of Wellington boots would be the best choice of styling product to recommend.  Hence in general styling products tend to be more conservative than hero products.  Our approach takes this difference into account by explicitly including this information as a feature. %\todo[inline]{Ben: How?}

% Method summary
%\todo[inline]{Ben: probably need a bit more here on our model}
We formulate our training task as binary classification, where GORDN learns to tell the difference between BTL and randomly generated negative outfits.  We consider an outfit to be a set of fashion items and train a model that projects items into a single \emph{style space}. Compatible items will appear close in style space enabling good outfits to be constructed from nearby items. GORDN is a neural network which combines embeddings of multi-modal features for all items in an outfit and outputs a single score.  When generating outfits, GORDN is used as a scorer to assess the validity of different combinations of items.

% Our contributions
In summary, our contributions are:
\begin{enumerate}
    \item A novel model that uses multi-modal data to generate outfits that can be trained on images in the wild i.e. dressed people rather than individual item flat shots. Outfits generated by our model outperform a challenging baseline by 21\% for womenswear and 34\% for menswear.
    \item A new research dataset consisting of 586,320 fashion outfits (images and textual descriptions) composed by ASOS stylists.  This is the world's largest annotated outfit dataset and is the first to contain Menswear items.
    %\item \todo[inline]{is this more general than other work? Oufits of many different types?}
\end{enumerate}

\section{Related work}
Our work follows an emerging body of related work on learning clothing style~\cite{Veit2015, Ma2017}, clothing compatibility~\cite{Shih2017, Song2018, Veit2015} and outfit composition~\cite{Vasileva2018, Li2017, Han2017, Hu2015}.  Successful outfit composition encompasses an understanding of both style and compatibility.

% Multi-modal embeddings
A popular approach is to embed items in a latent \emph{style} or \emph{compatibility} space often using multi-modal features~\cite{Song2018, Veit2015, Li2017, Tangseng2017}. A challenge with this approach is how to use item embeddings to measure the overall outfit compatibility. This challenge is increased when considering outfits of multiple sizes. Song et al.~\cite{Song2018} only consider outfits of size 2 made of top-bottom pairs. Veit et al.~\cite{Veit2015} use a Siamese CNN, a technique which allows only consideration of pairwise compatibilities.  Li et al.~\cite{Li2017} combine text and image embeddings to create multi-modal item embeddings which are then combined using pooling to create an overall outfit representation.  Pooling allows them to consider outfits of variable size. Tangseng et al.~\cite{Tangseng2017} create item embeddings solely from images. They are able to use outfits of variable size by padding their set of item images to a fixed length with a `mean image'. Our method is similar to these as we combine multi-modal item embeddings, however we aim not to lose information by pooling or padding.

% Vasileva
Vasileva et al.~\cite{Vasileva2018} extend this concept by noting that compatibility is dependent on context - in this case the pair of clothing types being matched.  They create learned type-aware projections from their style space to calculate compatibility between different types of clothing.  

\section{Outfit Datasets}
% Is this data going to be useful to the research community.  What is in it and how reproducable
% Reference description of embedding production to be in methodology section

\begin{table*}[t]
  \caption{Statistics of the ASOS outfits dataset}
  \label{tab:dataset}
  \begin{tabular}{*{7}{c}}
    \toprule
    Department&Number of Outfits&Number of Items&Outfits of size 2&Outfits of size 3&Outfits of size 4&Outfits of size 5\\
    \midrule
     % & Train & 237,478 & 239,818 & 108,096 & 85,906 & 36,328 & 7,148 \\
    %Womenswear & Test & 76,722 & 81,854 & 46,987 & 23,402 & 5,700 & 633 \\
      Womenswear & 314,200 & 321,672 & 155,083 & 109,308 & 42,028 & 7,781\\
     % & Train & 201,844 & 198,947 & 62,837 & 78,266 & 50,880 & 9,861 \\
    %Menswear & Test & 70,276 & 71,106 & 37,558 & 24,400 & 7,664 & 654 \\
      Menswear & 272,120 & 270,053 & 100,395 & 102,666 & 58,544 & 10,515\\
  \bottomrule
\end{tabular}
\end{table*}

% Buy the Look
The ASOS outfits dataset consists of 586,520 outfits, each containing between 2 and 5 items (see Table \ref{tab:dataset}). In total these outfits contain 591,725 unique items representing 18 different womenswear (WW) product types and 22 different menswear (MW) product types. As all of our outfits have been created by ASOS stylists, they are representative of a particular fashion style. 

Most previous outfit generators have used either co-purchase data from Amazon \cite{Veit2015, McAuley2015a} or user created outfits taken from Polyvore \cite{Vasileva2018, Han2017, He2018b, Song2018, Li2017, Tangseng2017, Nakamura2018}, both of which represent a diverse range of styles and tastes.  Co-purchase is not a strong signal of compatibility as co-purchased items are typically not bought with the intention of being worn together.  Instead it is more likely to reflect a user's style preference.  Data collected from Polyvore gives a stronger signal of compatibility and furthermore provide complete outfits.
% After we know which subset of outfits we will be using, we can analyse number of items, product types etc and perhaps compare with Polyvore datasets. 

% Comparison to Polyvore dataset. 
The largest previously available outfits dataset was collected from Polyvore and contained 68,306 outfits and 365,054 items entirely from WW~\cite{Vasileva2018}.  Our dataset is the first to contain MW as well. Our WW dataset contains an order of magnitude more outfits than the Polyvore set, but has slightly fewer fashion items.  This is a consequence of ASOS stylists choosing styling products from a subset of items held in our studios meaning that styling products can appear in many outfits. %The fact that many of our fashion items appear in multiple outfits is an advantage, as it means GORDN can learn that some items can be compatible with a diverse range of other items.

% Overview of distribution of MW/WW, high level product types and categories for hero and styling products separately to show the differences

\begin{figure}
    \centering
    \includegraphics[width=0.46\textwidth]{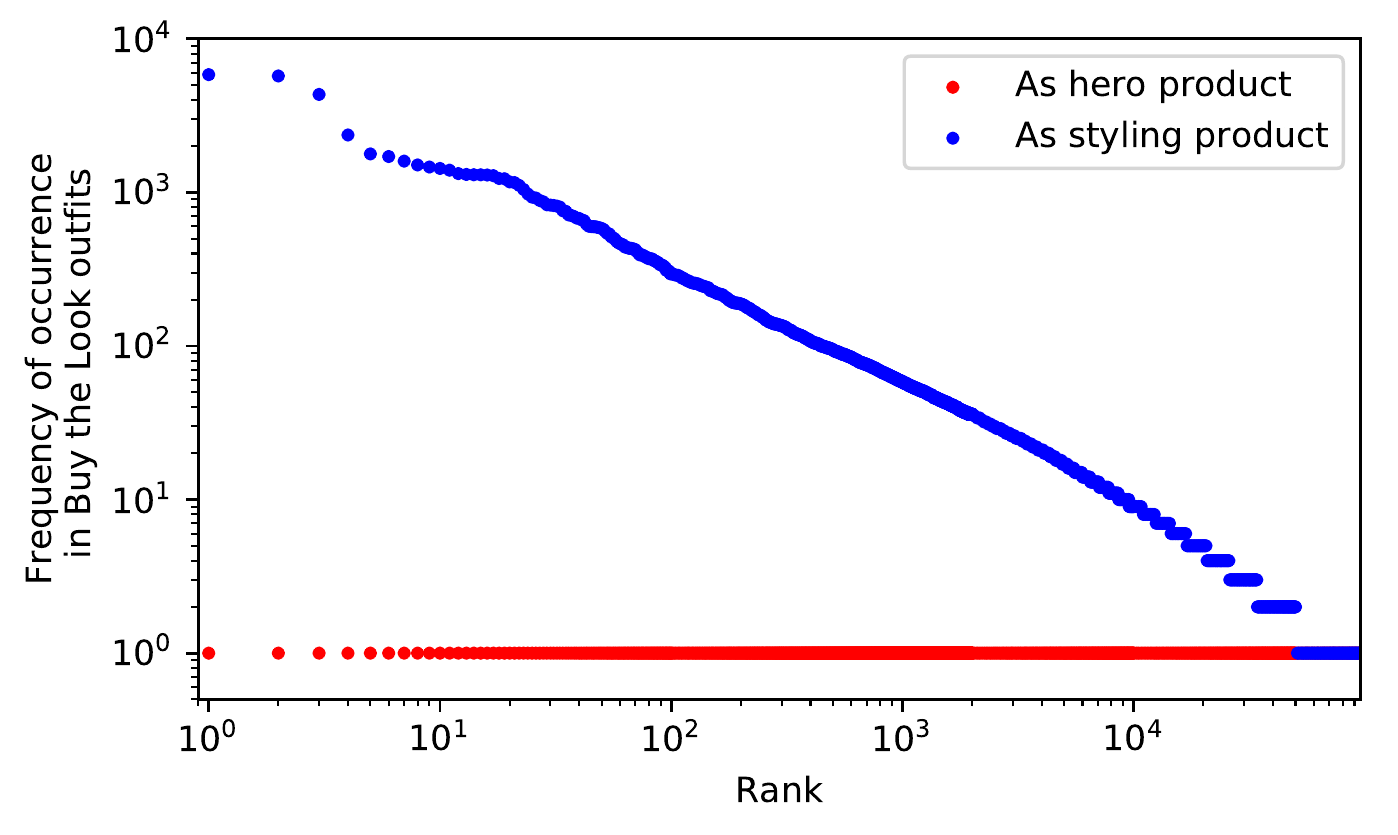}
    \caption{Frequency of occurrence of each womenswear item in our ASOS outfits dataset. Items are ranked by how frequently they occur as styling products. Each item appears once at most as a hero product (red), while there is a heavily skewed distribution in the frequency with which items appear as styling products (blue).}
    \label{fig:stylingdist}
    \vspace{-5mm}
\end{figure}

% Data available as input for each product
For each item we have four images, a text title and description, a high-level product type and a product category. We process both the images and the text title and description to obtain lower-dimensional embeddings, which are included in this dataset alongside the raw images and text to allow full reproducibility of our work. The methods used to extract these embeddings are described in Sections \ref{ssec:visualfeatures} and \ref{ssec:textfeatures},  respectively. Although we have four images for each item, in these experiments we only use the first image as it consistently shows the entire item, from the front, within the context of an outfit, whilst the other images can focus on close ups or different angles, and do not follow consistent rules between product types.
% Elaine: cut description of images?

\section{Methodology}
% How we treat outfits in this paper and overview of embedding approach
Our approach uses a deep neural network.  We acknowledge some recent approaches that use LSTM neural networks \cite{Han2017, Nakamura2018}.  We have not adopted this approach because fundamentally an outfit is a set of fashion items and treating it as a sequence is an artificial construct. Furthermore LSTMs are designed to progressively forget past items when moving through a sequence which in this context would mean that compatibility is not enforced between all outfit items. 

We consider an outfit to be a set of fashion items of arbitrary length which match stylistically and can be worn together. In order for the outfit to work, each item must be compatible with all other items. Our aim is to model this by embedding each item into a latent space such that for two items $({I_i}, {I_j})$ the dot product of their embeddings $(\textbf{z}_i, \textbf{z}_j)$ reflects their compatibility. We aim for the embeddings of compatible items to have large dot products and the embeddings of items which are incompatible to have small dot products. We map input data for each item $I_i$ to its embedding $\vec{z}_i$ via a multi-layer neural network. As we are treating hero products and styling products differently, we learn two embeddings in the same space for each item; one for when the item is the hero product, $\vec{z}_i^{(h)}$ and one for when the item is a styling product, $\vec{z}_i^{(s)}$; which is reminiscent of the context specific representations in language modelling \cite{Mikolov2013, Pennington2014a}.

\subsection{Network Architecture}
\begin{figure}
    \includegraphics[width=0.48\textwidth]{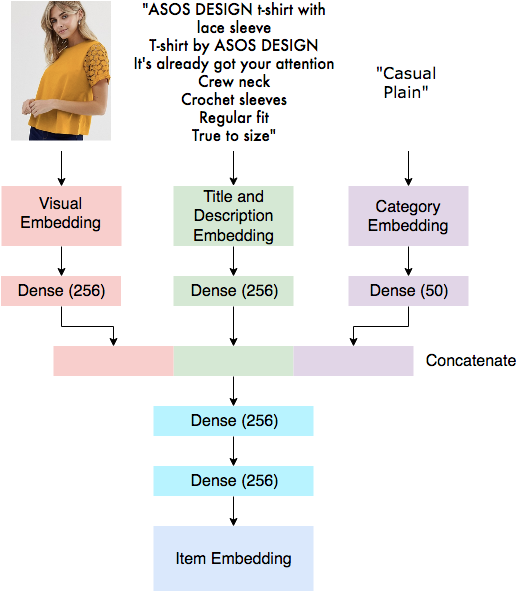}
    \caption{Network architecture of GORDN's item embedder. For each item the embedder takes visual features, a textual embedding of the item's title and description, a pre-trained GloVe embedding of the item's product category and a binary flag indicating if the item is the outfit's hero product. Each set of features is passed through a dense layer and the outputs of these layers are concatenated along with the hero product flag before being passed through two further dense layers. The output is an embedding for the item in our \emph{style} space. We train separate item embedders for womenswear and menswear items.}
    \label{fig:itemembedder}
    \vspace{-2mm}
\end{figure}

% description of network that produces product embedding - need to add notation for various inputs
For each item, the inputs to our network are a textual title and description embedding (1024 dimensions), a visual embedding (512 dimensions), a pre-trained GloVe embedding \cite{Pennington2014a} for each product category (50 dimensions) and a binary flag indicating the hero product. First, each of the three input feature vectors is passed through their own fully connected ReLU layer. The outputs from these layers, as well as the hero product flag, are then concatenated and passed through two further fully connected ReLU layers to produce an item embedding with 256 dimensions (Figure \ref{fig:itemembedder}). We use batch normalization after each fully connected layer and a dropout rate of 0.5 during training. 

\subsection{Outfit Scoring}
\begin{figure}
    \includegraphics[width=0.48\textwidth]{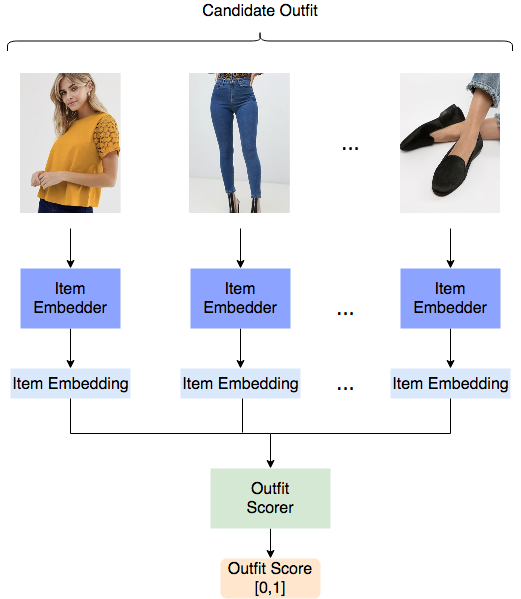}
    \caption{GORDN's outfit scorer takes the learnt embeddings of each item in an outfit and produces a score in the range [0,1], with a high value representing a compatible outfit. The scorer takes the sum of the compatibility scores (dot product) for each pair of items within the outfit, normalised by the number of pairs of items. This is then passed through a sigmoid function.}
    \label{fig:outfitscorer}
    \vspace{-2mm}
\end{figure}

% outfit scoring function
We use the dot product of item embeddings to quantify pairwise compatibility.  Outfit compatibility is then calculated as the sum over pairwise dot products for all pairs of items in the outfit (Figure \ref{fig:outfitscorer}).

For an outfit $\mathcal{S} = \{I_1, I_2, ... , I_N\}$ consisting of \emph{N} items, the overall outfit score is defined by
\begin{equation}
    y(\mathcal{S}) = \sigma\left(\frac{1}{N(N-1)}\sum_{\substack{i,j=1\\i<j}}^{N} \textbf{z}_i \cdotp \textbf{z}_j \right),
\end{equation}
where $\sigma$ is the sigmoid function.  The normalisation factor of $N(N-1)$, proportional to the number of pairs of items in the outfit is required to deal with outfits containing varying numbers of items.  The sigmoid function is used to ensure the output is in the range [0,1].

\subsection{Visual Feature Extraction}
\label{ssec:visualfeatures}
As described in Section \ref{sec:intro} and illustrated in Figure \ref{fig:buythelook}, items are photographed as part of an outfit and therefore our item images frequently contain the other items from the BTL outfit. Feeding the whole image to the network would result in features capturing information for the entire input leaking information to GORDN. It was therefore necessary to localise the target item within the image. To extract visual features from the images in our dataset we use VGG \cite{Simonyan14c}. Feeding the whole image to the network would result in features capturing information for the entire input, that is both the hero and the styling products. To extract features focused on the most relevant areas of the image, we adopt an approach based on Class Activation Mapping (CAM) \cite{zhou2016cvpr}. Weakly-supervised object localisation is performed by calculating a heatmap (CAM) from the feature maps of the last convolutional layer of a CNN, which highlights the discriminative regions in the input used for image classification. The CAM is calculated as a linear combination of the feature maps weighted by the corresponding class weights.

Before using the CAM model to extract image features, we fine-tune it on our dataset. Similar to~\cite{zhou2016cvpr} our model architecture combines VGG with a Global Average Pooling (GAP) layer and an output classification layer. We initialize VGG with weights pre-trained on ImageNet and fine-tune it towards product type classification (e.g. Jeans, Dresses, etc.). After training we pass each image to the VGG and obtain the feature maps.

\begin{figure*}
    \centering
    \includegraphics[width=0.6\textwidth]{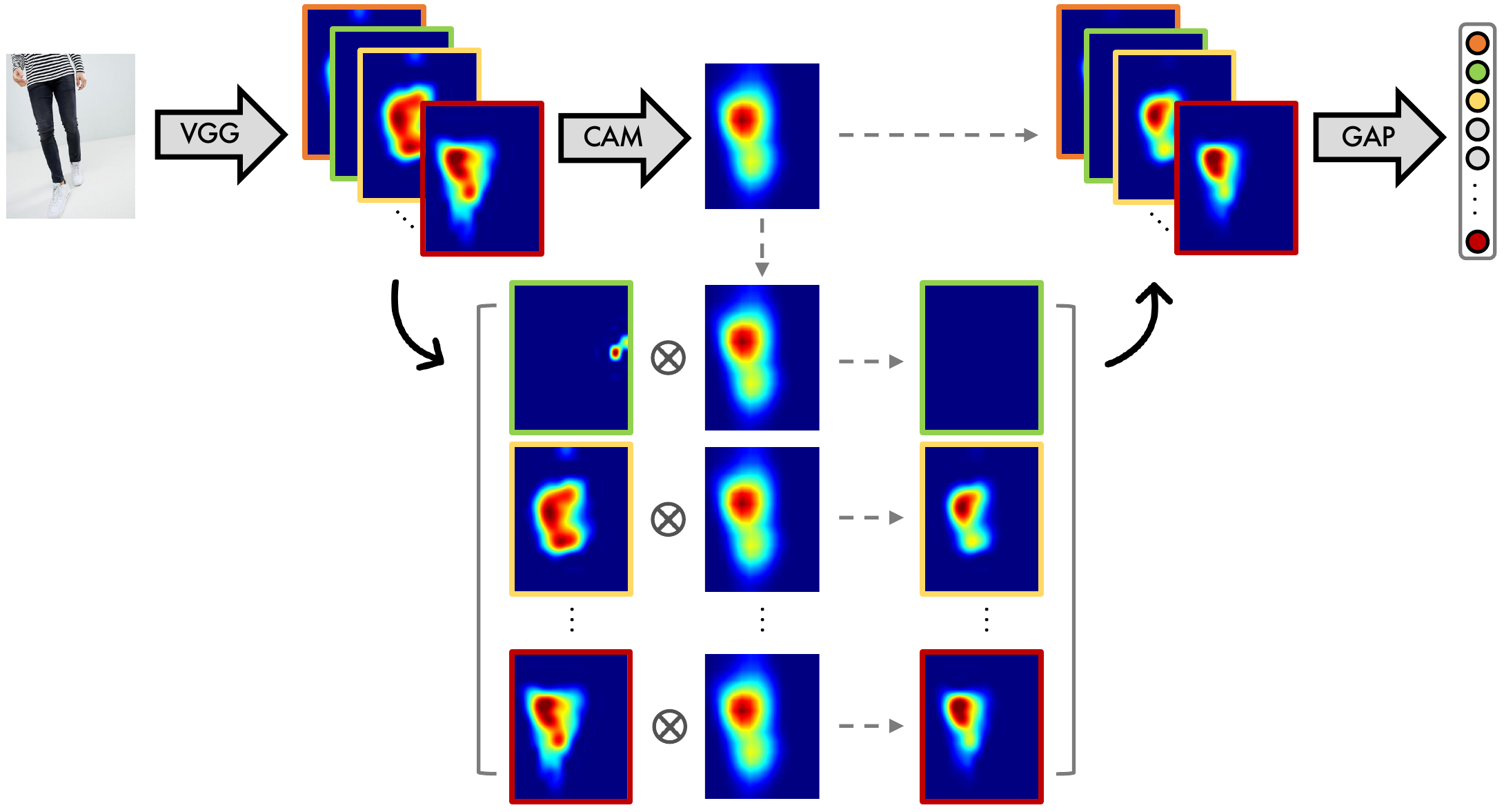}
    \caption{Pipeline for extracting image embeddings. An image is passed to VGG and the final convolutional feature maps are used to calculate the class activation map (CAM). The CAM is then used to spatially re-weight the feature maps by element-wise multiplication. Finally, a global average pooling (GAP) layer averages each re-weighted feature map to calculate a single value (shown in same colours) and outputs a 512-dimensional image embedding. During training, re-weighting is ignored and the output of the model is passed into a fully-connected layer for product type classification. We can see the effect of feature map re-weighting in the brackets for the case of an item with trousers as the hero product and top and shoes as styling products. Activations in the feature maps that correspond to the relevant region of interest in the input image (trousers) are refined by re-weighting (i.e. second and third row) whereas irrelevant activations are ignored (first row).}
    \label{fig:cam_pipeline_and_features}
\end{figure*}

To produce localised image embeddings, we use the CAM to spatially re-weight the feature maps. Similar to Jimenez et al.~\cite{Jimenez_2017_BMVC}, we perform the re-weighting by a simple spatial element-wise multiplication of the feature maps with the CAM. Our pipeline is shown in Figure \ref{fig:cam_pipeline_and_features}. This re-weighting can be seen as a form of attention mechanism on the area of interest in the image. The final image embedding is a 512-dimensional vector. The same figure illustrates the effect of the re-weighting mechanism on the feature maps.

% use a single figure for CAM-features, ignore the two bellow
\begin{comment}
    \begin{figure}
        \centering
        \includegraphics[width=0.48\textwidth]{figures/cam.pdf}
        \caption{Pipeline for re-weighting the image features. An image is passed to VGG and the final convolutional feature maps are used to produced the class activation map (CAM). The CAM is then used to spatially re-weight the feature maps by element-wise multiplication. Finally, a global average pooling (GAP) layer calculates a single value for each feature map (shown in same colours) and outputs a 512-dimensional image embedding. During training, this vector is passed to an output classification layer.}
        \label{fig:cam}
    \end{figure}
    
    \begin{figure}
        \centering
        \includegraphics[width=0.48\textwidth]{figures/features_reweighted.pdf}
        \caption{Re-weighting the image features on three different items. In the first row we see the product image along with their corresponding CAM (and overlaid). In the second row there are examples of convolutional feature maps before re-weighting with the CAM. In the last row there are the same maps after re-weighting. We can see how activations in the relevant regions of interest are refined by re-weighting (e.g. first product, first column) and irrelevant activations are ignored (e.g. third product, third column).}
        \label{fig:feature_maps_reweighted}
    \end{figure}
\end{comment}

\subsection{Title and Description Embeddings}
\label{ssec:textfeatures}
Product titles typically contain important information, such as the brand and colour. Similarly, our text descriptions contain details such as the item's fit, design and material. We use pre-trained text embeddings of our item's title and description. These embeddings are learned as part of an existing ASOS production system that predicts product attributes~\cite{Cardoso2018a}. Vector representations for each word are passed through a simple 1D convolutional layer, followed by a max-over-time pooling layer and finally a dense layer, resulting in 1024 dimensional embeddings.

\subsection{Training}
% binary classification and negative sample generation
We train GORDN in a supervised manner using a binary cross-entropy loss. Our training data consists of positive outfit samples taken from the ASOS outfits dataset and randomly generated negative outfit samples. We generate negative samples for our training and test sets by randomly replacing the styling products in each outfit with another item of the same type. For example, for an outfit with a top as the hero product and jeans and shoes as styling products, we would create a negative sample by replacing the jeans and shoes with randomly sampled jeans and shoes. We ensure that styling products appear with the same frequency in the positive and negative samples by sampling styling products from their distribution in the positive samples.  This is important as the frequency distribution of styling products is heavily skewed (Figure~\ref{fig:stylingdist}) and without preserving this GORDN could memorise frequently occurring items and predict outfit compatibility based on their presence.  By matching the distribution GORDN must instead learn the characteristics of items which lead to compatibility. Although some of the negative outfits generated in this way may be good quality outfits, we assume that the majority of these randomly generated outfits will contain incompatible item combinations. Randomly selecting negative samples in this way is common practice in metric learning and ranking problems (e.g. \cite{Rendle2012, Hoffer2015}). In both training and testing, we generate one negative outfit sample for each positive outfit sample.

To assess the relative importance of each set of input features, we conduct an ablation study. We separately train five different versions of GORDN using only the textual title and description embeddings as input (text), only the visual embeddings (vis), both text and visual embeddings (text + vis), text, visual and category embeddings (text + vis + cat), and finally the full set of inputs (text + vis + cat + hero). For each of these configurations we trained 20 models from scratch using Adam~\cite{Kingma2015} for 30 epochs.

\subsection{Outfit Generation Method}
\label{sec:beamsearch}
% generating outfit based on template
Once trained, we use GORDN to generate novel outfits.  GORDN can generate outfits of any length by sequentially adding items and re-scoring the new outfit. Each outfit starts with a hero product from our catalogue. We then define an outfit template $ \mathcal{P} = \{T^{(h)}, T_1, ..., T_{N-1}\}$ as a set of product types including the hero product type $T^{(h)}$ and $N-1$ other compatible styling product types. Our aim is to find the set of items of the appropriate product types that maximises the outfit score \emph{y}. 

\begin{figure}
    \centering
    \includegraphics[width=0.48\textwidth, trim=4mm 0 0 1mm]{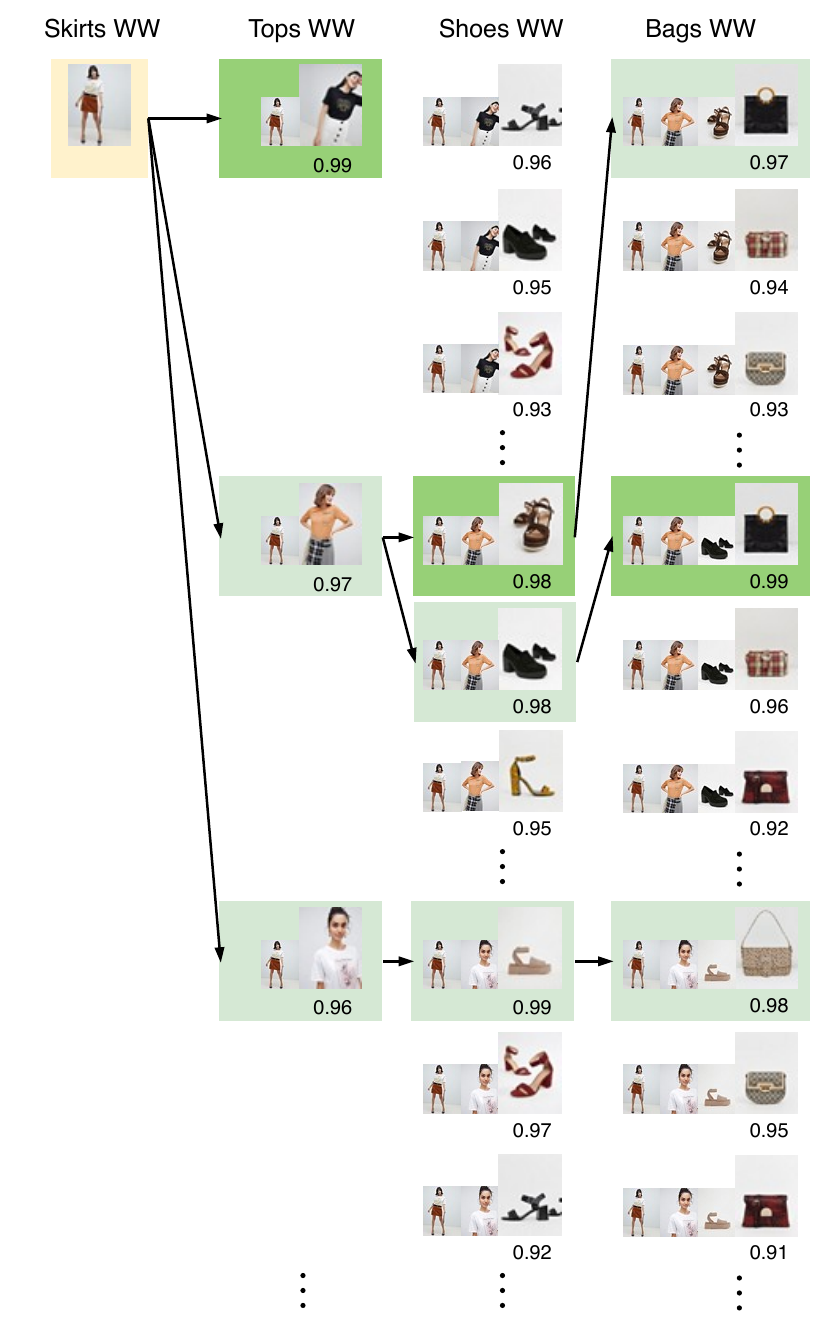}
    \caption{Beam search in the context of outfit generation. Starting with an outfit template of product types and a hero product (highlighted in yellow) each product type in the template is filled sequentially by finding the products from the catalogue which when added to the outfit give the highest outfit score.  After each step the number of outfits retained is reduced to the beam width (set to 3).  The retained outfits after each step are highlighted in green with the highest scoring outfit in dark green.}
    \label{fig:beamsearch}
    \vspace{-2mm}
\end{figure}

An exhaustive search of every possible combination of styling products is combinatorial over the number of product types and cannot be computed within a reasonable e-commerce latency budget we use a beam search to complete the outfit (illustrated in Figure \ref{fig:beamsearch}). We use a beam width of 3 for all our outfit generation as we found this returned the optimal outfit 77.5\% of the time.  The beam search algorithm is repeated for all $(N-1)!$ permutations of the styling product types in the template as different outfits may be generated depending on the order in which product types are added.  The final outfit returned is the one that has maximal score across all template permutations.

%We start with a candidate outfit containing just the hero product, select a product type from $\mathcal{P}$ and find the top $w$ items of the required product type that maximise the outfit score. The results of this make up $w$ candidate sets each containing two items. For each candidate set we then pick another product type from $\mathcal{P}$ and repeat the procedure.  This generates $w^2$ candidate sets which are then cut down to the top $w$.  The process is repeated until all product types in $\mathcal{P}$ are filled. The procedure is performed for all permutations of the outfit template as different outfits may be generated depending on the order product types are added. The final outfit returned is the one that has maximal score across all template permutations. $w$ is known as the \emph{beam width} and in our experiment we set $w=3$ which we found returned the optimal outfit 77.5\% of the time.  

% how was template selected for AB test - most popular
The choice of template $\mathcal{P}$ depends on the use case. Templates for each hero product type can be found from our ASOS outfits dataset. The distribution of templates can be used to introduce variety into the generated outfits. For the purposes of our AB test, we picked the most frequently occurring template for each hero product type. % table showing outfit templates used in AB test?

\section{Evaluation}

We evaluate the performance of GORDN on two tasks. The first task is binary classification of genuine and randomly generated outfits, using a held out test set. The second task is user evaluation of outfits generated by GORDN in comparison to randomly generated outfits from a simple baseline model.

\subsection{Train/test split}
\label{sec:traintestsplit}
% Train/test split

\begin{table}[t]
  \caption{The number of outfits and items in our training and test partitions after applying the Louvain community detection method to the full ASOS outfits dataset. There are no items which appear in both the training and test set.}
  \label{tab:traintestsplit}
  \begin{tabular}{*{4}{c}}
    \toprule
    Department&Dataset&Number of Outfits&Number of Items\\
    \midrule
    Womenswear & Train & 237,478 & 239,818 \\
    & Test & 76,722 & 81,854 \\
    \midrule
    Menswear & Train & 201,844 & 198,947 \\
    & Test & 70,276 & 71,106 \\
  \bottomrule
\end{tabular}
\end{table}

% TODO: More detail about train test split both methodology and numbers
We split the ASOS outfits dataset first into WW and MW and each of these into a training and test set ensuring that no items appeared in both sets. To achieve this we first represented the ASOS outfits dataset as a graph where the nodes are items and edge weights are defined by the number of outfits pairs of items are found together in. We then used the Louvain community detection method \cite{Blondel2008b} to split the graph into communities which maximise the modularity.  This resulted in many small communities which could then be combined together to create the train and test sets.  When re-combining communities care was taken firstly to respect the desired train-test split ratio as far as possible and secondly to ensure items from each season are proportionally split between the train and test sets. This resulted in 76:24 and 74:26 train-test splits in terms of outfits for WW and MW respectively. The use of disjoint train and test sets provides a sterner test for GORDN as it is unable to simply memorise which items frequently co-occur in outfits in the training set. Instead, the embeddings GORDN learns must represent product attributes that contribute to fashion compatibility. 

\begin{table}[t]
    \centering
    \caption{Comparison of GORDN when using different features on the binary classification task. Scores are the mean over 20 runs of the test set AUC after 30 epochs of training.}
    \begin{tabular}{l c c}
    \toprule
    Features & \multicolumn{2}{c}{AUC} \\
    & WW & MW \\
    \midrule
    vis & 0.66 & 0.55 \\ % ebgpuvm01
    text & 0.80 & 0.66 \\ % max: 0.80 & 0.66 ebgpuvm01
    text + vis & 0.82 & 0.66 \\ % max: 0.82 & 0.67 ebgpuvm01
    text + vis + cat & 0.82 & 0.67 \\ % max: 0.827 & 0.681 ebgpuvm02
    text + vis + cat + hero & \textbf{0.83} & \textbf{0.67} \\ % max: \textbf{0.8318} & \textbf{0.6834} Stephen
    \bottomrule
    \end{tabular}
    \vspace{-2mm}
    \label{tab:classificationresults}
\end{table}

\subsection{Outfit Classification Results}
The test set contains BTL outfits and an equal number of negative samples.  We use GORDN to predict compatibility scores for the test set outfits and then calculate the AUC of the ROC curve.  We found that training separate versions of GORDN for WW and MW produced better results and so we report the performance of these here. 

% Describe our results
Table \ref{tab:classificationresults} shows the AUC scores achieved for different combinations of features.  As we add features to GORDN we increase its performance, with the best performing model including text, visual, category and hero item features. The majority of the performance benefit came from the text embeddings with visual embeddings adding a small improvement.  We expected our visual embeddings to be of poorer quality than those for Polyvore datasets as our images show whole outfits on people as opposed to a photograph of the fashion item in isolation.  In contrast the success of our text embeddings could be due to the attribution task on which they were trained \cite{Cardoso2018a}.  A total of 34 attributes were predicted, including many attributes that are directly applicable for outfit composition e.g. `pattern', `neckline', `dress type' and `shirt style'. 

% Why WW scores are better than MW
For all feature combinations the WW model greatly outperforms the MW one.  This could be due to fashion items being more interchangeable in MW than in WW hence having more similar embeddings making the training task harder.  For example the mean correlations between the text embeddings for the most prevalent product type in the WW and MW training sets are 0.041 (dresses) and 0.077 (T-shirts) respectively.  More simply, there are many combinations of MW T-shirts and jeans that make equally acceptable outfits whereas there are far fewer for WW dresses and shoes.

% Compare to other people's results
Using GORDN to predict compatibility scores for our test set is equivalent to the \emph{outfit compatibility} task used by \cite{Han2017} and \cite{Vasileva2018}.  As noted by Vasileva et al., Han's negative samples contain outfits that are incompatible due to multiple occurrences of product types e.g. multiple pairs of shoes in the same outfit. Since our negative samples were generated using templates respecting product type our data does not have this characteristic and hence we compare only to results in \cite{Vasileva2018}.  Our WW model achieves an AUC score just slightly less than Vasileva et al.'s compatibility AUC on their disjoint Polyvore outfits dataset.  
% We achieve similar AUC scores to Vasileva - with the caveat that this is a completely different dataset.  Our reliance on images is much less than other papers due to the poor quality of our images.  We make up for this with our text embeddings.

% Could also show results for outfits of different sizes and discuss the differences in these

\subsection{Generated Outfit Evaluation}
% This section should include
% - How we generated the outfits (model + random) for the test
% - Description of internal app + picture
% - Results calculation

% Data generation
We perform an AB test to evaluate the quality of outfits generated by GORDN. We select six popular outfit templates to test, three each for WW and MW (shown in Table \ref{tab:abtestresults}), and generate 100 WW and 100 MW outfits split evenly across the templates. We use a large pool of in stock products from which we randomly select hero products of the required product types.  The remaining items in the outfits were generated using the beam search method described in Section \ref{sec:beamsearch} and illustrated in Figure \ref{fig:beamsearch}.  These outfits constitute our test group.  For a control group we take the same hero products and templates and generate outfits by randomly selecting items of the correct type from the same pool of products. By using outfit templates we ensure that none of the outfits contain incompatible product type combinations, such as by pairing a dress and a skirt, or by placing two pairs of shoes in one outfit. Instead, the quality of the outfits depends solely on style compatibility between items.

\begin{figure}
    \includegraphics[width=0.48\textwidth]{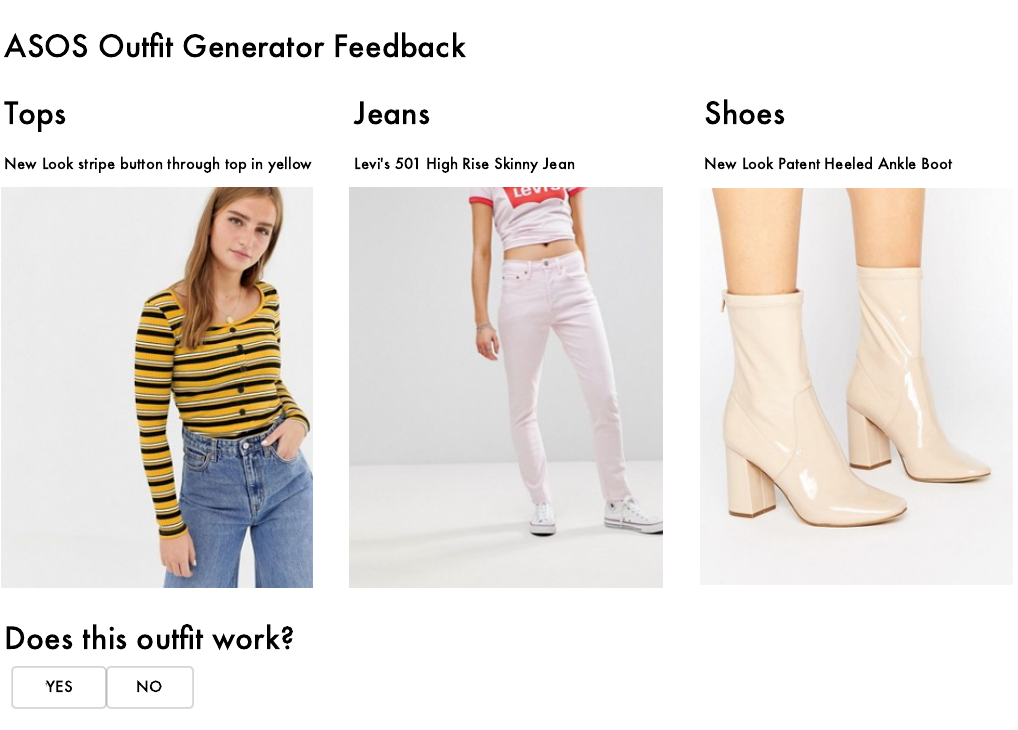}
    \caption{Screenshot of Outfit Evaluation App}
    \label{fig:feedbackapp}
    \vspace{-2mm}
\end{figure}

% App description
To run the AB test we developed an internal app which we exposed to ASOS employees. A screenshot of the app is shown in Figure \ref{fig:feedbackapp}. The app displayed an outfit to the user asking them to decide if the items in the outfit work stylistically. The outfits were shown one at a time to each user with the order of outfits randomised for each user.  WW and MW outfits were only shown to female and male users respectively and each user rated all 200 outfits from their corresponding gender. 

\begin{figure}
    \centering
    \includegraphics[width=0.46\textwidth]{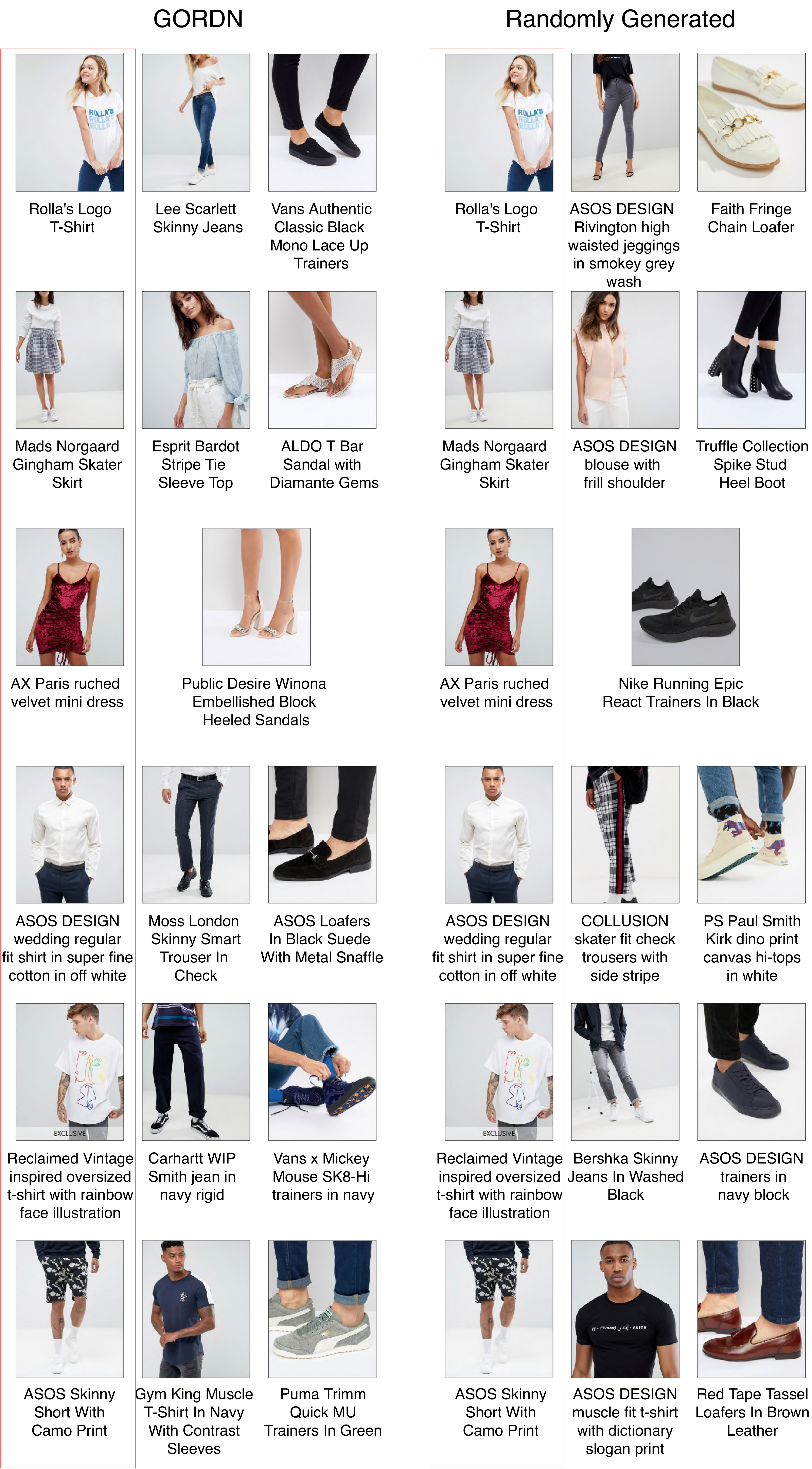}
    \caption{Example outfits generated by GORDN (left column) and by random selection of items (right column) that were used in our AB test. In each row the same hero product is used (red box) and each model is given the same template of product types.}
    \label{fig:exampleoutfits}
    \vspace{-2mm}
\end{figure}

% Data & results calculation
The data collected from the app comprised a binary score for each user-outfit pair. The data exhibit two way correlation --- all scores from the same user are correlated due to the inherent user preferences and all scores on the same outfit are also correlated. We therefore used a two-way random effects model as described in \cite{Ribeiro2011} to calculate the variance of the sample mean. We could then use a t-test for the difference between means to calculate if the difference between the test and control groups was significant.

% Results
The results are shown in Figure \ref{tab:abtestresults}.  We analyse the results for WW and MW separately as the WW and MW models were trained separately. We collected 1,200 observations per group for WW and 900 for MW. We found the relative difference between the test and control groups to be 21.28\% and 34.16\% for WW and MW respectively. Testing at the 1\% level these differences were significant. We were able to further break down our results to find that GORDN outperformed the control significantly for all templates.

\begin{table*}[t]
    \centering
    \caption{Relative differences between the test and control group user scores. All results are significant at the 1\% level.}
    \begin{tabular}{l l c c c c}
    \toprule
    & & Ctrl score & Test score & Rel. diff. (\%) & p-value \\
    \midrule
    WW & all & 0.49 & 0.60 & 21.28 & < 0.01 \\
    & Dress | Shoes & 0.54 & 0.78 & 46.12 & < 0.01 \\
    & Tops | Jeans | Shoes & 0.61 & 0.64 & 4.53 & < 0.01 \\
    & Skirts | Tops | Shoes & 0.33 & 0.36 & 10.77 & < 0.01 \\
    MW & all & 0.49 & 0.66 & 34.16 & < 0.01 \\
    & T-Shirts | Jeans | Shoes, Boots \& Trainers & 0.63 & 0.76 & 19.07 & < 0.01 \\
    & Shirts | Trousers \& Chinos | Shoes, Boots \& Trainers & 0.42 & 0.60 & 44.35 & < 0.01 \\
    & Shorts | T-Shirts | Shoes, Boots \& Trainers & 0.43 & 0.63 & 47.24 & < 0.01 \\
    \bottomrule
    \end{tabular}
    \label{tab:abtestresults}
\end{table*}

% Results discussion
% Mention easier to create good outfits by random for some templates eg MW t-shirts, jeans, trainers?

% Example outfits
Examples of outfits generated for our AB test are shown in Figure \ref{fig:exampleoutfits}. For each hero product we show the outfit produced by GORDN alongside the randomly generated outfit. Many of the random examples appear to be reasonable outfits. Although the random model is simple, the use of outfit templates, combined with selecting only products that were in stock in the ASOS catalogue on the same day makes this a challenging baseline.

\subsection{Style space}
\begin{figure*}
    \centering
    \begin{subfigure}{0.98\textwidth}
        \centering
        \includegraphics[width=0.79\textwidth]{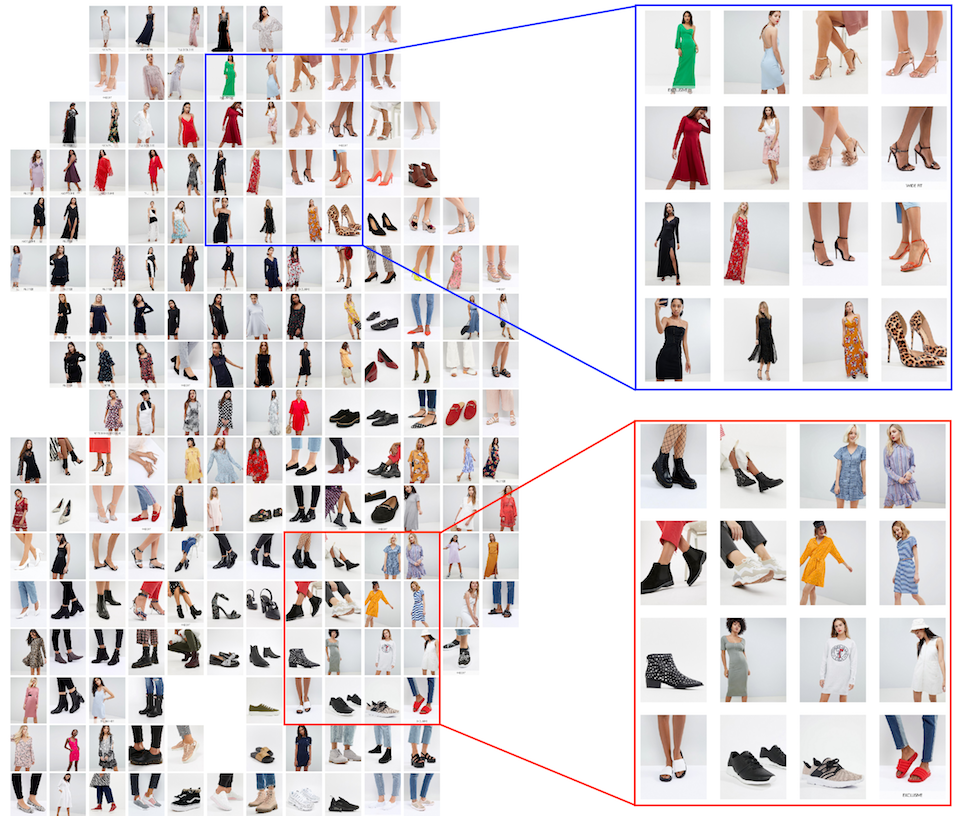}
        \caption{}
        \label{fig:tsne_a}
    \end{subfigure}
    \begin{subfigure}{0.98\textwidth}
        \centering
        \includegraphics[width=0.85\textwidth]{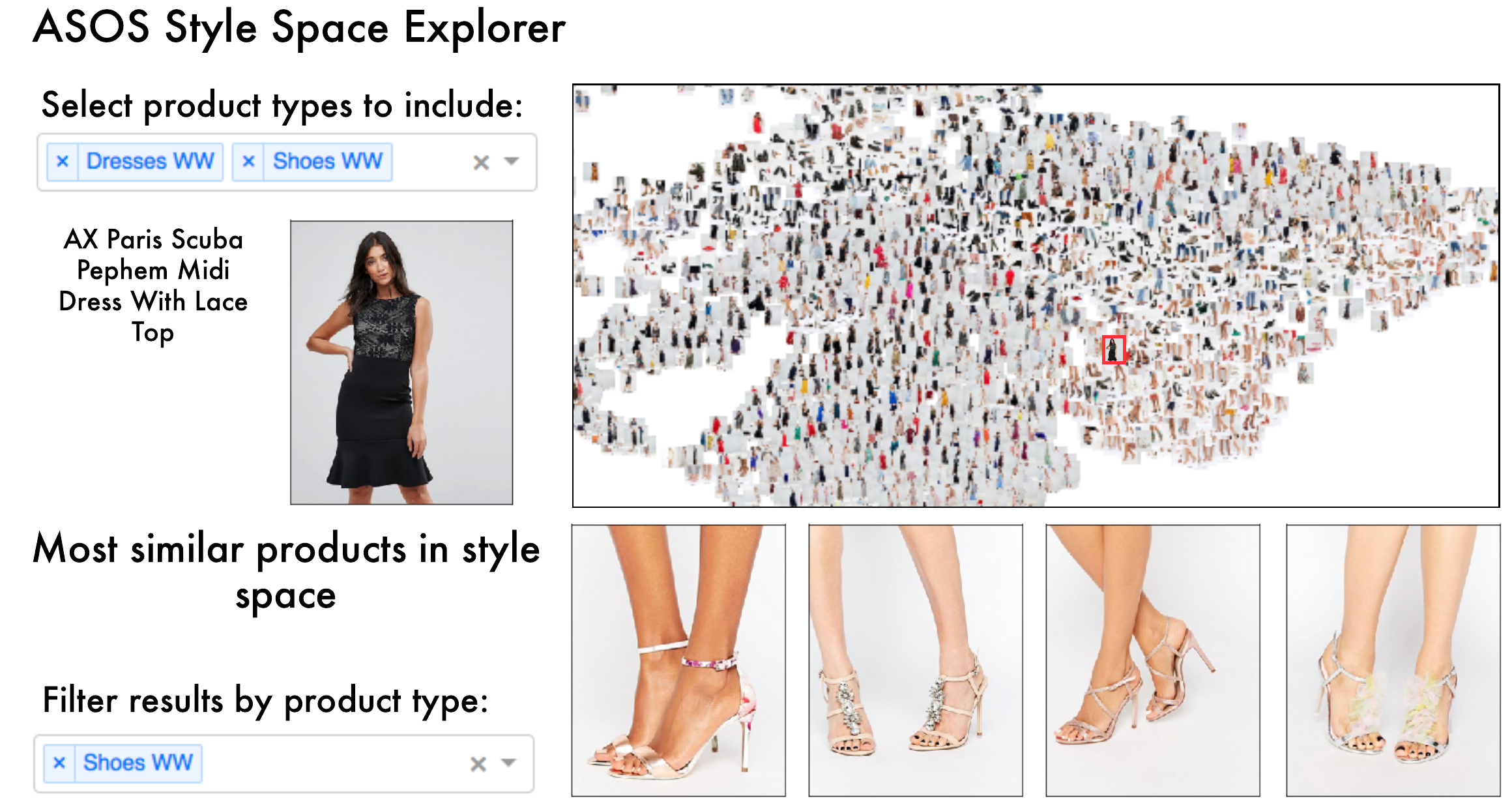}
        \caption{}
        \label{fig:styleexplorer}
    \end{subfigure}
    \caption{a) A section of a t-Distributed Stochastic Neighbour Embedding (t-SNE) visualisation of the embeddings learnt by GORDN for womenswear dresses and shoes. Similar items have similar embeddings, but so do compatible items of different types. The two highlighted areas illustrate that casual dresses are embedded close to casual shoes (red), while occasion dresses are embedded close to occasion shoes (blue). b) Screenshot of an internal app, developed to allow exploration of the learnt style space. Users can create t-SNE plots using different product types and then select individual items to view the most similar items in style space of different types.}
    \label{fig:tsne}
\end{figure*}

% Brief description of our t-SNE plots and discussion of how compatible items are close to one another in style space
We visualise our style space using a t-Distributed Stochastic Neighbour Embedding (t-SNE)~\cite{vanderMaaten2008} plot in two dimensions (Figure~\ref{fig:tsne_a}). While similar items have similar embeddings, we can also see that compatible items of different product types have similar embeddings. Rather than dresses and shoes being completely separate in style space, these product types overlap, with casual dresses having similar embeddings to casual shoes and occasion dresses having similar embeddings to occasion shoes. We built an app for internal use that uses t-SNE to visualise our style space and allows us to easily explore compatible item combinations, as predicted by GORDN (Figure~\ref{fig:styleexplorer}).

%\afterpage{\FloatBarrier}

\section{Conclusion}

We have described GORDN, a multi-modal neural network for generating outfits of fashion items, currently under development at ASOS. GORDN learns to represent items in a latent style space, such that compatible items of different types have similar embeddings. GORDN is trained on the ASOS outfits dataset, a new resource for the research community which contains over 500,000 outfits curated by professional stylists. The results of an AB test show that users approve of outfits generated by GORDN 21\% and 34\% more frequently than those generated by a simple baseline model for womenswear and menswear, respectively.

%\end{document}  % This is where a 'short' article might terminate

%\begin{acks}
%  The authors would like to thank Dr. Yuhua Li for providing the
%  MATLAB code of the \textit{BEPS} method.

%  The authors would also like to thank the anonymous referees for
%  their valuable comments and helpful suggestions. The work is
%  supported by the \grantsponsor{GS501100001809}{National Natural
%    Science Foundation of
%    China}{http://dx.doi.org/10.13039/501100001809} under Grant
%  No.:~\grantnum{GS501100001809}{61273304}
%  and~\grantnum[http://www.nnsf.cn/youngscientists]{GS501100001809}{Young
%    Scientists' Support Program}.

%\end{acks}

\bibliographystyle{ACM-Reference-Format}
\newcommand{\showDOI}[1]{\unskip}
\newcommand{\showURL}[1]{\unskip}
\renewcommand{\showeprint}[8]{\unskip}
\bibliography{outfit_recs}

\end{document}